\documentclass[preprint,epsfig,onecolumn,floats, showkeys]{revtex4}
\usepackage{makeidx}
\usepackage{amsmath}
\usepackage{amssymb}
\usepackage{graphicx}
\usepackage{epsfig}

\begin{document}

\title{Applying Local Clustering Method to Improve the Running Speed of Ant
Colony Optimization}
\author{Chao-Yang Pang}
\email{cypang@live.com}
\email{cypang@sicnu.edu.cn}
\author{Wei Hu}
\author{Xia Li }
\author{Ben-Qiong Hu}

\begin{abstract}
Ant Colony Optimization (ACO) has time complexity $O(tmN^2)$, and its
typical application is to solve Traveling Salesman Problem (TSP), where $t$,
$m$, and $N$ denotes the iteration number, number of ants, number of cities
respectively. Cutting down running time is one of study focuses, and one way
is to decrease parameter $t$ and $N$, especially $N$. For this focus, the
following method is presented in this paper. Firstly, design a novel
clustering algorithm named Special Local Clustering algorithm (SLC), then
apply it to classify all cities into compact classes, where compact class is
the class that all cities in this class cluster tightly in a small region.
Secondly, let ACO act on every class to get a local TSP route. Thirdly, all
local TSP routes are jointed to form solution. Fourthly, the inaccuracy of
solution caused by clustering is eliminated. Simulation shows that the
presented method improves the running speed of ACO by 200 factors at least.
And this high speed is benefit from two factors. One is that class has small
size and parameter $N$ is cut down. The route length at every iteration step
is convergent when ACO acts on compact class. The other factor is that,
using the convergence of route length as termination criterion of ACO and
parameter $t$ is cut down
\end{abstract}

\maketitle

\address{Key Lab. of Visual Computation and Virtual Reality of Sichuan
Province, Sichuan Normal University, Chengdu 610068, China} \address{College
of Mathematics and Software Science, Sichuan Normal University, Chengdu
610068, China}

\address{College of Mathematics and Software Science, Sichuan Normal
University, Chengdu 610068, China}

\address{College of Information Engineering, Shenzhen University, Shenzhen,
Guangdong province 518060, China }

\address{College of Information Management, Chengdu University of
Technology, Chengdu 610059, China}

\section{INTRODUCTION}

\subsection{Introduction of Ant Colony Optimization (ACO)}

In 1991, Ant Colony Optimization (ACO) was presented firstly by M. Drigo\cite%
{A. Colorni11}and applied to solve TSP firstly by V. Mahiezzo et al.\cite{A.
Colorni11,M Dorigo12,M Dorigo13}. Drigo et al. create a new research topic
which is studied by many scholars now.

ACO is essentially a system based on agents that simulate the natural
behavior of ants, in which real ants are able to find the shortest route
from a food source to their nest, without using visual cues by exploiting
pheromone information\cite{M Dorigo13}. Pheromone is deposited when ants
walking on a route. It provides heuristic information for other ants to
choose their routes. The more dense the pheromone trail of a route is, the
more possibly the route is selected by ants. At last, nearly all ants select
the route that has the most dense pheromone trail, and it's the shortest
route potentially.

ACO has been applied to solve optimization problems widely and successfully,
such as TSP\cite{A. Colorni11,M Dorigo12,M Dorigo13,Hai-Bin Duan14},
quadratic assignment problem\cite{Maniezzo20}, image processing\cite{S.
Meshoul24}, data mining\cite{Rafael26}, classification or clustering analysis%
\cite{X. Li28}, biology\cite{P. Meksangsouy29}. The application of ACO leads
the theoretic study of ACO. Gutijahr firstly analyzes the convergence
property of ACO\cite{W. J. Gutijahr36}. Stutzle \& Dorigo prove the
important conclusion, that if the running time of ACO is long enough, ACO
can find optimal solution possibly\cite{T Stuezle37}. The other interesting
property is revealed currently by Birattari and et al. that the sequence of
solutions of some algorithms does not depend on the scale of problem instance%
\cite{Mauro38}.

Running time is too long and the quality of solution is still low, that are
the two main problems of ACO. To solve the main problems, the configuration
of the parameters is discussed\cite{M Dorigo12,M Dorigo13}. The method of
adaptation is used to improve ACO\cite{M Dorigo39}. Parallel computation and
other method are used to accelerate ACO\cite{Bullnheimer40}.

\subsection{Clustering Correlates to the Running Time of ACO}

One of study focus of ACO is to cut down running time. The running time of
ACO is $O(t_{\max }MN^{2})$, and $M=[N/1.5]$ in general, where $t_{\max }$, $%
M$ and $N$ denote the iteration number, number of ants, and number of cities
respectively\cite{Hai-Bin Duan14}. The running time is proportional to $%
N^{2} $. Cutting down the number of cities $N$ is the key to reduce running
time. Therefore, classifying all cities into different classes and letting
ACO act on each class will reduce running time heavily. Hu and Huang used
this method to improve the running speed of ACO\cite{Xiao bing HU52}, which
named ACO-K-Means. It's faster than ACO by factors of 5-15 approximately.
Simulations show that ACO-K-Means algorithm is valid only to the set of
cities that has evident clustering feature, and invalid to more general
situation. ACO-K-Means implies that it is possible that using clustering
method to improve the running speed of ACO.

\subsection{Introduction of Local Clustering Algorithm}

Clustering is the classification of objects of a set (named training set)
into different classes (or groups), so that the data in each class (ideally)
share some common trait. One of most popular clustering algorithm is K-Means
Clustering Algorithm\cite{Y. Linde47,Chao-Yang Pang48}. K-Means Clustering
Algorithm assigns each point to the cluster whose center (i.e., centroid) is
nearest to it, then update the centroid. Repeat this process until
termination criterion is satisfied \cite{Chao-Yang Pang48}.

During the $t-th$ iteration of K-Means algorithm, the $i-th$ class has
distortion that is defined as the average distance of each point and the
class centroid, which is denoted by $D_{i}^{(t)}$ $(1\leq i\leq m)$, where $%
m $ is the number of classes. Pang proves that for each $i$, the distortion
sequence $\{D_{i}^{(t)}\}$ is convergent if the $i-th$ class is separated
from other classes evidently \cite{Chao-Yang Pang48}. That is, distortion
sequence is convergent locally. According to this property, an algorithm
named \textbf{local clustering algorithm (LC)} is presented\cite{Chao-Yang
Pang49}, its essential idea is introduced as below.

Step1. K-Means is applied to a given training set to generate classes.

Step2. The class which distortion $D_{i}^{(t)}$ is convergent firstly is
deleted from training set. Then update training set such that it is
comprised by residual points. Go to step1.

Stop the process of step1-2 until all data is classified.

LC algorithm is faster than K-Means algorithm by factors of 4-13
approximately.

Suppose the $i-th$ class is $R_{i}^{(t)}$ during the $t-th$ iteration of
K-Means algorithm. Set $R_{i}^{(t)}$ has entropy $H(R_{i}^{(t)})$, where $%
H(R_{i}^{(t)})=\underset{a\in R_{i}^{(t)}}{-\sum }p(a)\log _{2}p(a)$ and $%
p(a)$ is the probability of data $a$. It is proved that entropy sequence $%
\{H(R_{i}^{(0)}),H(R_{i}^{(1)}),\cdots ,H(R_{i}^{(t)}),\cdots \}$ is
convergent \cite{Chao-Yang Pang48}. That is, the convergent criterion of
K-Means algorithm can be replaced by the convergence of entropy sequence
\cite{Chao-Yang Pang50}. The K-Means with convergent criterion of entropy
convergence is fast by two of factors at least \cite{Chao-Yang Pang50, Li
X51}.

\section{IMPROVE LOCAL CLUSTERING ALGORITHM TO GENERATE COMPACT CLASS}

\subsection{Compact Set and The Method of Generation}

For any subset of Euclidean space $R^{n}$, every sequence in this subset has
a convergent subsequence, the limit point of which belongs to the set. This
subset is called \textbf{compact set}. The conception of compact set (or
compactness) is a topology conception. To understand it easily, compactness
can be described visually as the phenomenon that many points cluster tightly
in a small region, while non-compact set is the set that most of points
clustering loosely in a big region.

K-Means Clustering, LC or other algorithm aim to partition a training set
into classes. Some classes are compact and some are not. The most common
situation is that a class contains a compact subset and some loose points,
and the compact subset is around the center of the class. That is, the
central part of class is compact possibly. To extract compact subset from a
class, the following $3\delta $-principle is introduced.

For Gauss distribution, suppose that $\delta $ denotes the deviation of
random data. It is the $3\delta $-principle that it's more than 99\%
probability that a random point falling into the central region of data set
which radius is $3\delta $ \cite{Chao-Yang Pang48}. The central region
contains points more than 99\%. Thus, if radius $3\delta $ is small enough
and the number of points is big enough, the central region is compact. Even
the central region with radius $3\delta $ is not compact, shortening the
radius of central region to $\frac{3\delta }{4}$,$\ \frac{3\delta }{16}$,
and so on will make it compact. For Gauss distribution which is comprised by
enough points, the compact central region always exists. In general, for a
class generated by clustering algorithm, all distances of points and class
centroid comprise a gauss distribution approximatively. Therefore, the
central region of a class is compact possibly.

Suppose the $i-th$ class is $R_{i}^{(t)}$ at the $t-th$ iteration of K-Means
or LC algorithm. With the increase of iteration,\ class sequence $%
\{R_{i}^{(0)},R_{i}^{(1)},\cdots R_{i}^{(t)},R_{i}^{(t+1)}\cdots \}$ $(1\leq
i\leq m)$ appears, where $m$ denotes the number of classes. Let%
\begin{equation}
D_{i}^{(t)}=\frac{1}{\left\vert R_{i}^{(t)}\right\vert }\underset{x\epsilon
R_{i}^{(t)}}{\sum }d(x,c_{i}^{(t)})  \label{dis21}
\end{equation}

, where $\left\vert R_{i}^{(t)}\right\vert $ denotes the number of elements
in $R_{i}^{(t)}$ and $d(x,c_{i}^{(t)})$ denotes distance.

\begin{equation}
\delta _{i}^{(t)}=\frac{1}{\left\vert R_{i}^{(t)}\right\vert }\underset{%
x\epsilon R_{i}^{(t)}}{\sum }\left\vert
d(x,c_{i}^{(t)})-D_{i}^{(t)}\right\vert  \label{de22}
\end{equation}

Clearly, $D_{i}^{(t)}$ is the distortion of class $R_{i}^{(t)}$ and $\delta
_{i}^{(t)}$ is the approximation of deviation of $D_{i}^{(t)}$.

\begin{equation}
K_{i}^{(t)}=\{x\mid d(x,c_{i}^{(t)})\leq \frac{1}{4^{p}}(D_{i}^{(t)}+3\delta
_{i}^{(t)}),x\epsilon R_{i}^{(t)}(p\geq 0)\}  \label{K23}
\end{equation}

\begin{figure}[tbh]
\begin{center}
\epsfig{file=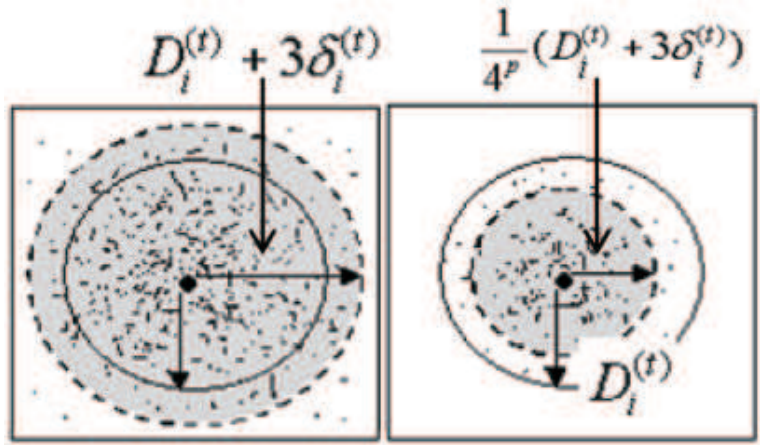,width=10cm,}
\end{center}
\caption{The illustration of compact central region of a class: In a class $%
R_{i}^{(t)}$, most of points cluster around their centroid and few points
are far away from the centroid. Subset $K_{i}^{(t)}$ (i.e., the shadow part)
is the central region of class $R_{i}^{(t)}$. Compact set is the set that
many points cluster in a small region tightly. Increasing parameter $p$ will
shorten radius and make $K_{i}^{(t)}$ compact.}
\label{fig1}
\end{figure}

$K_{i}^{(t)}$ is the central region of class $R_{i}^{(t)}$. Parameter $p$ is
used to shorten the radius of central region $K_{i}^{(t)}$ and makes it
compact. Fig.\ref{fig1} illustrates the $3\delta $-principle and compact
subset $K_{i}^{(t)}$.

\subsection{Subroutine 1: Local Clustering Algorithm with $3\protect\delta $%
-Principle}

The local clustering algorithm with $3\delta $-principle is used to classify
points into classes and to extract compact central region of classes. Its
essential idea is described as below.

Firstly, apply LC algorithm to cluster data. And apply the criterion of
entropy convergence (i.e.,$\frac{\left\vert
H(R_{i}^{(t)})-H(R_{i}^{(t+1)})\right\vert }{H(R_{i}^{(t)})}\rightarrow 0$)
to mark the stable class $R_{i}^{(t)}$.

Secondly, Extract compact central region $K_{i}^{(t)}$ from class $%
R_{i}^{(t)}$ and preserve it as a genuine class. Remove $R_{i}^{(t)}$ from
training set and update it. Repeat above two steps until all compact central
regions are extracted. The detail is described as below.

{\small Input parameters:}

$T${\small : Training Set}

$m${\small : The number of class}

$\varepsilon ${\small : The stop threshold for clustering.}

$C^{(0)}=\{c_{i}^{(0)}(1\leq i\leq m)${\small : Initial centroids set.}

$p${\small : A parameter to adjust the size of compact subsets }$%
K_{i}^{(t)}(p\geq 0)${\small .}

{\small Output:}

$\varphi (T)=\{K_{1}^{(t)},K_{2}^{(t)},\cdots ,K_{i}^{(t)},\cdots
K_{m}^{(t)}\}${\small \ (i.e., the set of compact subset, see fig.\ref{fig1})%
}

$\sigma (T)=\{B_{1}^{(t)},B_{2}^{(t)},\cdots ,B_{i}^{(t)},\cdots
B_{m}^{(t)}\}${\small , where }$B_{i}^{(t)}=R_{i}^{(t)}-K_{i}^{(t)}${\small %
, and it is comprised by dispersive points ( }$1\leq i\leq m${\small , see
fig.\ref{fig1})}

\textbf{void Subroutine1 ( }$T$\textbf{,}$m$\textbf{,}$\varepsilon $\textbf{,%
}$C^{(0)}$\textbf{,}$p$\textbf{,}$\varphi (T)$\textbf{,}$\sigma (T)$\textbf{)%
}

{\small \{}

{\small \ Step1. Initialization: Let iteration number }$t=0${\small . Let }$%
Counter=m${\small . Let }$\varphi (T)=\phi ${\small \ and }$\sigma (T)=\phi $%
{\small , where }$\phi ${\small \ denotes empty set. According to initial
centroids set }$C^{(0)}${\small , generate initial partition of training set
}$\varphi ^{(0)}=\{R_{i}^{(0)}\mid R_{i}^{(0)}\subset T,1\leq i\leq m\}$%
{\small .}

{\small \ Step2. While (}$Counter>0${\small ) \ \{}

{\small \ \ \ \ \ Step2.1. Generate new centroids set }$\
C^{(t+1)}=\{c_{i}^{(t+1)}\mid 1\leq i\leq m)\}${\small \ \ and new partition
}$\varphi ^{(t+1)}=\{R_{i}^{(t+1)}\mid 1\leq i\leq m\}$

{\small \ /* Note: Check whether entropy sequence }$%
\{H(R_{i}^{(0)}),H(R_{i}^{(1)}),\cdots ,H(R_{i}^{(t)}),\cdots \}${\small \
is convergent. If it is convergent, let the convergent marker }$%
StableMarker(R_{i}^{(t+1)})=True${\small */}

{\small \ \ \ \ \ Step2.2. For (}$i=1${\small ; }$i\leq Counter${\small ;
i++) \ \{}

{\small \ \ \ \ \ \ \ \ Estimate the entropy of class }$R_{i}^{(t+1)}$%
{\small , i.e., }$H(R_{i}^{(t+1)})=\log _{2}\left\vert
R_{i}^{(t+1)}\right\vert ${\small .}

{\small \ \ \ \ \ \ \ \ If (}$\frac{\left\vert
H(R_{i}^{(t)})-H(R_{i}^{(t+1)})\right\vert }{H(R_{i}^{(t)})}<\varepsilon $%
{\small ) \ \{ }$StableMarker(R_{i}^{(t+1)})=True${\small ; \}}

{\small \ \ \ \ \ \ \ \ else \{}$StableMarker(R_{i}^{(t+1)})=False${\small \}%
}

{\small \ \ \ \ \ \}}

{\small \ /*Note: Extract the data around the centroid of class as a genuine
class */}

{\small \ \ \ \ \ Step2.3. For (}$i=1${\small ; }$i\leq Counter${\small ;
i++) \ \{}

{\small \ \ \ \ \ \ \ \ If (}$StableMarker(R_{i}^{(t+1)})=True${\small ) \ \{%
}

{\small \ \ \ \ \ \ \ \ Calculate compact central region }$K_{i}^{(t)}$%
{\small \ according to formula (\ref{K23})}

{\small \ \ \ \ \ \ \ \ Calculate }$B_{i}^{(t)}${\small : }$%
B_{i}^{(t)}=R_{i}^{(t+1)}-K_{i}^{(t)}$

{\small \ \ \ \ \ \ \ \ Let }$\varphi (T)=\varphi (T)\cup K_{i}^{(t)}$

{\small \ \ \ \ \ \ \ \ Let }$\sigma (T)=\sigma (T)\cup B_{i}^{(t)}$

{\small \ \ \ \ \ \ \ \ Update Training Set: }$T=T-R_{i}^{(t)}$

{\small \ \ \ \ \ \ \ \ Update centroids set: }$C^{(t+1)}=C^{(t+1)}-%
\{c_{i}^{(t+1)}\}$

${\small \ \ \ \ \ \ \ \ Counter=Counter-1}$

{\small \ \ \ \ \ \}}

{\small \ \ \}}

$\ \ t=t+1${\small .}

{\small \ \}}

{\small \}}

\subsection{Special LC Algorithm to Generate Compact Classes (SLC)}

Note that above subroutine 1 is not a partition of training set. Subroutine
1 extracts only compact central regions of all classes and the residual
points are unclassified. The residual points comprise a new training set.
And it's possible that some of residual points cluster together tightly and
comprise some small compact subsets again. These small compact subsets are
new classes. To obtain these new classes and classify all points, SLC
algorithm is described as below.

{\small Input parameters:}

$T_{0}${\small : Training Set}

$m_{0}${\small : The initial number of classes.}

$\varepsilon ${\small : The stop threshold for clustering.}

{\small Output:}

$Num${\small : The final number of classes.}

$CLS${\small : The partition of }$T_{0}${\small , in which each class is
compact.}

{\small SLC Algorithm:}

{\small Step1. Initialization: Let }$T=T_{0}${\small , }$m=m_{0}${\small , }$%
CLS=\phi ${\small , and }$p=0${\small .}

{\small Step2. For (i=0;}$i<[\log _{2}m]${\small ; i++) \ /*Note: }$[\log
_{2}m]${\small \ denotes the integer*/}

{\small \{ Step2.1 Generate initial centroids set }$\
C^{(0)}=\{c_{i}^{(0)}\mid 1\leq i\leq m)\}${\small .}

{\small \ Step2.2 Call Subroutine1 ( }$T${\small ,}$m${\small ,}$\varepsilon
${\small ,}$C^{(0)}${\small ,}$p${\small ,}$\varphi (T)${\small ,}$\sigma
(T) ${\small )}

{\small \ Step2.3 }$CLS=CLS\cup \varphi (T)${\small ;}

{\small \ Step2.4 }$T=\sigma (T)${\small ;}

{\small \ /*Note: Increase }$p${\small \ to get smaller compact class*/}

{\small Step2.5 }$m=[\frac{m}{2}]${\small ; }$p=p+1${\small ;}

{\small \}}

{\small Step3 Every residual point }$x${\small \ in the last set }$\sigma
(T) ${\small \ is regarded as a class }$\{x\}${\small . And let }$%
CLS=CLS\cup \{x\}${\small . Let }$Num${\small \ denote the number of classes
contained in }$CLS${\small . The two outputs are }$CLS${\small \ and }$Num$%
{\small .}

\subsection{The Clustering for Mixture Distribution (SLC-Mixture)}

The clustering algorithm SLC presented above generates spherical classes
only. However, for a general distribution, some classes are spherical shape,
some classes are chain shape in which points cluster closely around a curve
(or a line), even some classes contain isolated points. This common
distribution is called \textbf{mixture distribution}. For a large-scale TSP,
the distribution of cities is mixture distribution in general. The
clustering method for mixture distribution is proposed as below.

\subsubsection{The Simple Maker to Distinguish Spherical Class from
Chain-Shaped Class}

\begin{figure}[tbh]
\begin{center}
\epsfig{file=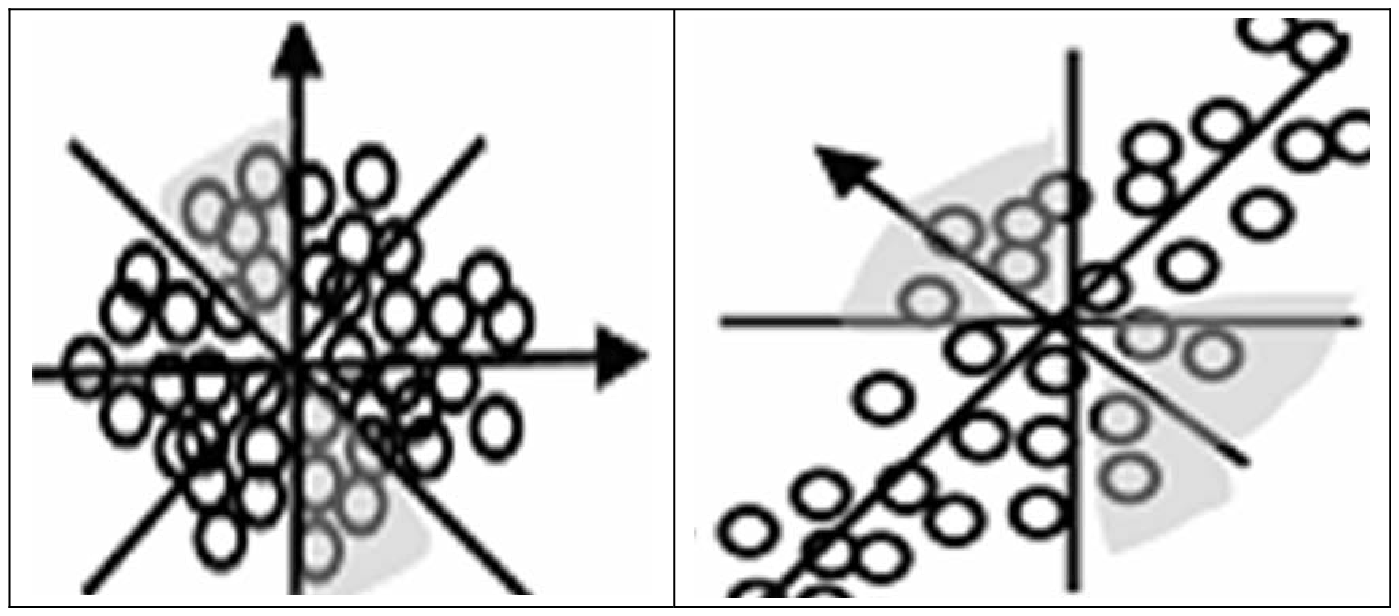,width=10cm,}
\end{center}
\caption{The Illustration of Distinguishing Spherical Class from
Chain-shaped Class: A given class is divide into 8 parts along the 8 lines
through the centroid of the class. If the class is spherical, the percentage
of each part is close to threshold $\protect\epsilon =1/8$. If the class is
chain-shaped class (or part of chain-shaped class), there are 2-4 parts
which percentage is far less than 1/8. Therefore, the percentage of each
part is the marker of spherical class. $\protect\varepsilon =0.058$ in this
paper because some classes are elliptical.}
\label{fig2}
\end{figure}

The position of city on a map is two-dimensional point. A given class can be
divided into 8 areas along the 4 directions of the north-south, west-east
and two diagonal lines through the centroid of the class. If the class is
spherical, the percentage of points in each area is close to 1/8 and is the
same approximately. If the class is chain-shaped class (or part of
chain-shaped class), it's impossible that the percentage of every area is
close to 1/8 at the same time. Therefore, the percentage of points in each
area is the maker of spherical class. Fig.\ref{fig2} illustrates the marker.

\subsubsection{Applying SLC to Process Mixture Distribution (SLC-Mixture)}

At first, apply SLC to classify all data of training set. Secondly, apply
the marker presented above to distinguish spherical classes and extract them
from the training set. Then all residual points comprise a new set named
residual set. The residual set contains only chain-shaped classes and
isolated points. Third, apply the method presented in Ref.\cite{HUANG
Xiaobin53} to classify all residual points of residual set into different
chain-shaped classes or marked as isolated points. The method presented in
Ref.\cite{HUANG Xiaobin53} is named Chain-Shaped Clustering Algorithm and is
introduced at appendix I.

The clustering\ method presented in this section is called \textbf{%
SLC-Mixture} algorithm, which processes the mixture distribution of
spherical classes, chain-shaped classes and isolated points.

\section{APPLY SLC TO ACO\qquad}

\subsection{The Termination Criterion of ACO}

Suppose ACO acts on a compact class and let $L_{t}$ denotes the minimum
route length that is generated at the $t-th$ iteration of computation. There
are sequence $\{L_{1},L_{2},\cdots ,L_{t},L_{t+1},\cdots \}$ and it is
convergent under idea condition. The convergent criterion $\frac{\left\vert
L_{t}-L_{t+1}\right\vert }{L_{t}}\leq \varepsilon $ is proposed as the
termination criterion of ACO in this paper.

In the following discussion, ACO refers to the algorithm which termination
criterion is $(\left\vert L_{t}-L_{t+1}\right\vert /L_{t})\leq \varepsilon $.

\subsection{Apply SLC to Improve The Running Speed of ACO (ACO-SLC)}

In this section, the clustering algorithm SLC will be applied to improve the
running speed of ACO. The method is named ACO-SLC and it is described as
below.

{\small Input Parameter:}

$T${\small : Set of cities}

{\small Output: \ The shortest TSP route obtained by the algorithm}

{\small ACO-SLC Algorithm:\qquad }

{\small Step1. Apply SLC algorithm to partition set }$T${\small . The
classes are }$B_{1},B_{2},\cdots ,B_{i},\cdots B_{Num}${\small , and their
centroids are }$b_{1},b_{2},\cdots ,b_{i},\cdots b_{Num}${\small \
respectively.}

{\small Stpe2. Construct graph }$G^{^{\prime }}${\small : Centroids }$%
b_{1},b_{2},\cdots ,b_{i},\cdots b_{Num}${\small \ are regarded as virtual
cities respectively, and the virtual cities are regarded as the vertices of
graph }$G^{^{\prime }}${\small . For a pair of classes }$B_{i}${\small \ and
}$B_{j}${\small , if there exists two cities that belong to }$B_{i}${\small %
\ and }$B_{j}${\small \ respectively and they joint each other, use an edge
to joint the two corresponding vertices }$b_{i}${\small \ and }$b_{j}$%
{\small . The weight of edge is the minimum distance between two classes,
i.e.,}

\begin{equation}
{\small d(B}_{i}{\small ,B}_{j}{\small )=}\min {\small \{d(x}_{i}{\small ,x}%
_{j}{\small )\mid x}_{i}{\small \epsilon B}_{i}{\small ,x}_{j}{\small %
\epsilon B}_{j}{\small \}}  \label{di}
\end{equation}

{\small Step3. Calculate a TSP route of graph }$G^{^{\prime }}${\small \ to
generate the traveling order of all classes: Let ACO algorithm act on graph }%
$G^{^{\prime }}${\small \ to find a TSP route denoted by }$%
b_{j_{1}},b_{j_{2}},\cdots b_{j_{Num}}${\small , where }$j_{1},j_{2},\cdots
j_{Num}${\small , is a permutation of sequence }$1,2,\cdots Num${\small .
The pair of classes }$B_{j_{i}}${\small \ and }$B_{j_{(i+1)}}${\small \ is
called neighbor class.}

{\small Step4. Choose an edge as the bridge to joint a pair of neighbor
classes, and this edge is named bridge edge: Assume that the two neighbor
classes are }$B_{j_{1}}${\small \ and }$B_{j_{2}}${\small . If there exist
an edge such that}

\begin{equation}
{\small d(x}_{u}{\small ,x}_{v}{\small )=}\min {\small \{d(a,b)\mid
a\epsilon B}_{j_{1}}{\small ,b\epsilon B}_{j_{2}}{\small \}}  \label{di2}
\end{equation}

{\small , edge }$(x_{u},x_{v})${\small \ is the bridge edge, }$x_{u}${\small %
\ and }$x_{v}${\small \ are called border cities, where vertices }$a${\small %
\ and }$b${\small \ should be not used to joint other neighbor classes.}

{\small Step5. Calculate a local TSP route for every class }$B_{i}(1\leq
i\leq Num)${\small : Add a new edge to joint the two border cities in the
class, and mark the edge as necessary edge of the local TSP route. This edge
is named pseudo-edge. Let the ACO algorithm with convergence criterion }$%
(\left\vert L_{t}-L_{t+1}\right\vert /L_{t})\leq \varepsilon ${\small \ act
on the class to generate a local TSP route.}

{\small Step6. Construct a TSP route: Walk along the traveling order
obtained at step3, for every pair of neighbor classes, delete the
pseudo-edge of each class such that the local route is not close. Then let
the local route of each class and the bridge edge between these two classes
be jointed.}

\begin{figure}[tbh]
\begin{center}
\epsfig{file=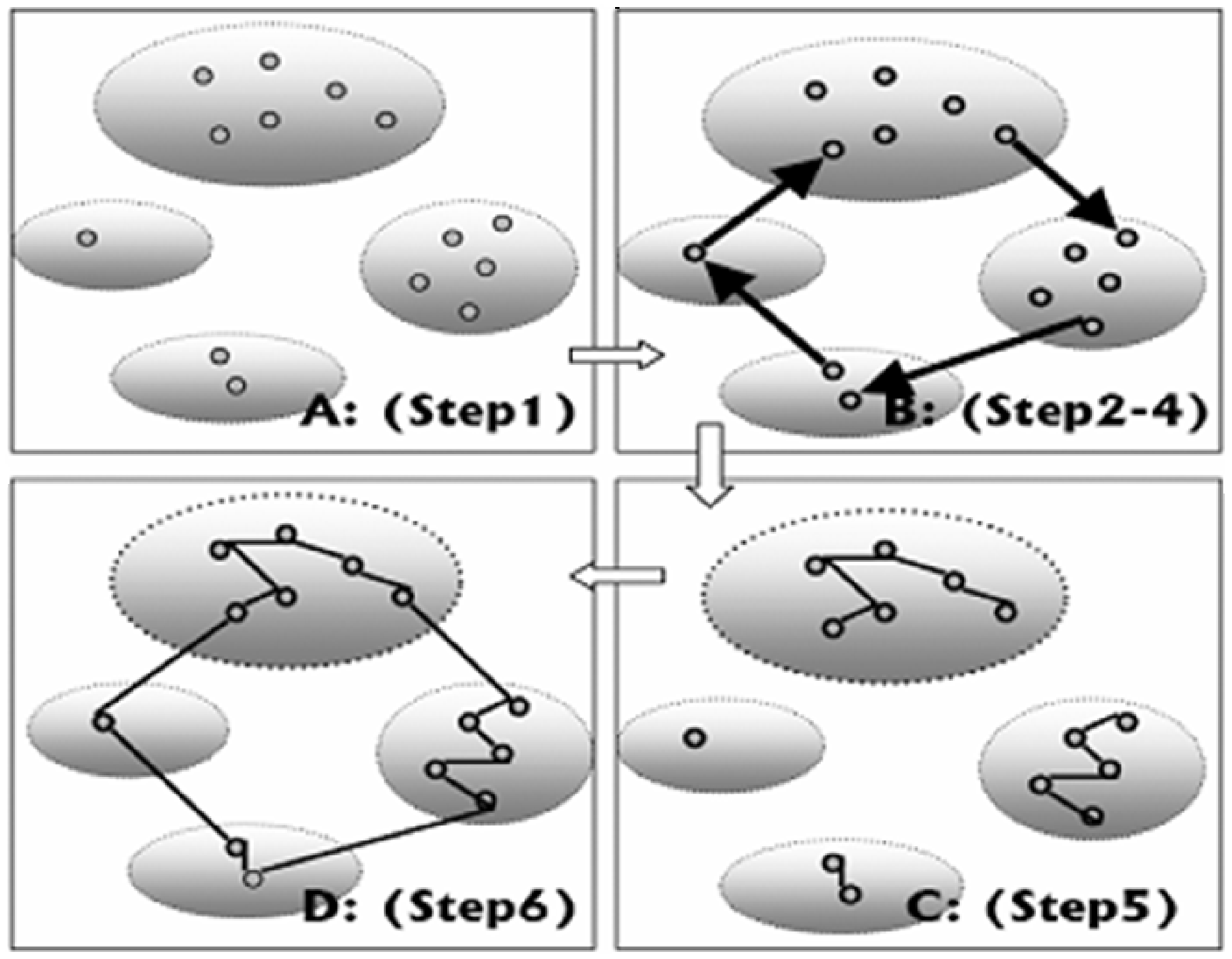,width=10cm,}
\end{center}
\caption{The schematic diagram of ACO-SLC : Firstly, classify all points
into compact classes. Secondly, the centroid of each class is regarded as a
virtual city, and calculate a virtual TSP route. Then along the virtual
route, joint all classes. Thirdly, let ACO act on each class to get a local
TSP route. Fourthly, joint all local TSP routes along the virtual route to
form the last route.}
\label{fig3}
\end{figure}

Fig.\ref{fig3} illustrates the processing of ACO-SLC algorithm.

\subsection{Using the Method of Little-window and Removing Cross-edge to
Improve ACO-SLC (ACO-SLC-LWCR)}

Clustering may cause the error of solution although it improves the running
speed of ACO heavily. If all classes are compact and separated clearly, the
quality of solution of ACO-SLC should be very good. However, in fact, the
border between two neighbor classes is fuzzy. The fuzzy border will cause
the inaccuracy of solution, and much longer route will appear. And
recognizing the longer part and removing it will generate better solution
possibly. It is well known that, the shortest route is always at the surface
of a convex hull. Thus, the longer part should be at the inner of a convex
hull and two longer edges intersect. In other words, intersection of two
edges is a marker of longer part of a route possibly. According to the
marker, removing longer edges is call \textbf{removing cross-edge} or
\textbf{removing intersection edges}, which is similar to the method in \cite%
{Hai-Bin Duan14}. ({\small Notice: in Ref.\cite{Hai-Bin Duan14}, before
execute ACO, the long and crossed edges are removed to improve the running
speed of ACO, not to improve the solution quality.})

\begin{figure}[tbh]
\begin{center}
\epsfig{file=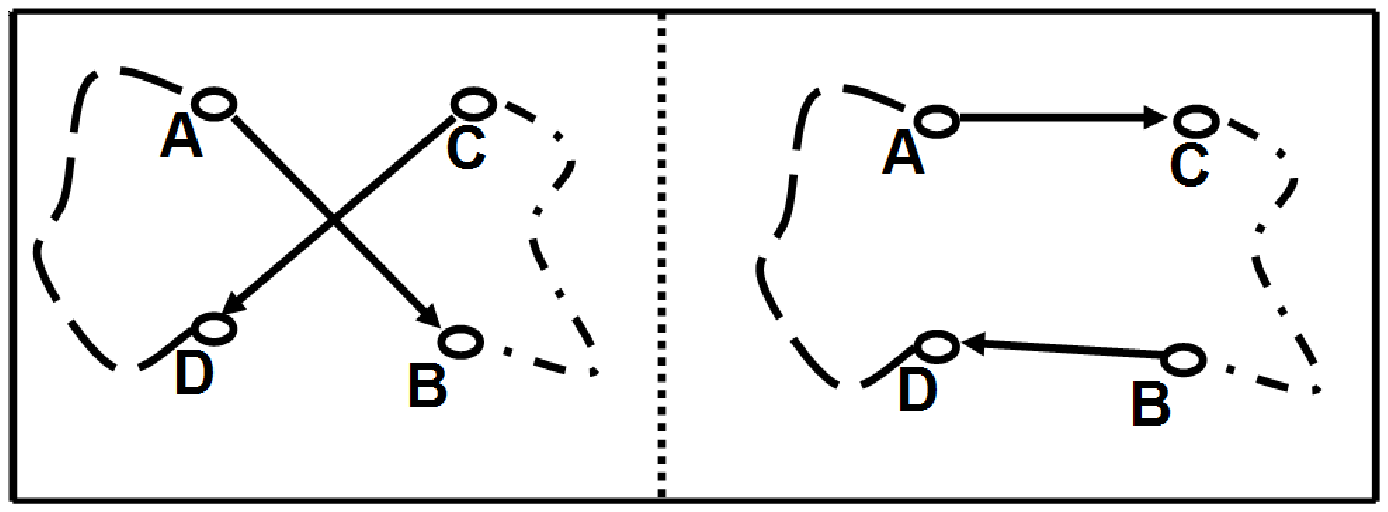,width=10cm,}
\end{center}
\caption{\textbf{The Illustration of Removing Cross-edges from TSP Route:}
At the left figure, $AB$ and $CD$ intersect each other. There is a principle
that shortest route is at the surface of a convex hull. Thus, edge $AB$ and $%
CD$ is the longer part of route and should be removed. Removing these two
edges will generate shorter route (see right figure). }
\label{fig4}
\end{figure}

Fig.\ref{fig4} illustrates the method of removing cross-edge.

In addition, a simple method named \textbf{little-window strategy} is
proposed to improve the running speed of ACO in Ref.\cite{XIAO46}. Construct
a set $S_{i}$ that is comprised by $w$ accessible and short edges which join
the $i-th$ city, where $w$ is a pre-assigned constant. The ant which has
arrived at $i-th$ city will select an edge from window set $S_{i}$ only to
arrive its next city, and not selection an edge from all neighbor edges of
this vertex. So, this method improve the running speed of ACO. In addition,
the parameter configuration of the method is put at appendix II.

The ACO-SLC with little-window strategy and cross-edges removing is called
\textbf{ACO-SLC-LWCR}.

\subsection{The ACO-SLC for Mixture Distribution (ACO-SLC-Mixture)}

ACO-SLC is suitable for the spherical shape distribution only, and the low
quality of solution will appear possibly when ACO-SLC is applied to process
mixture distribution. To process mixture distribution, the following method
named ACO-SLC-Mixture is proposed in this paper.

Firstly, apply SLC-Mixture at section 2.4.2 to partition the set of cities
into spherical classes, chain-shaped classes, or isolated points. Secondly,
apply ACO-SLC-LWCR to each class and generate a TSP route.

\section{SIMULATION}

In this section, five related algorithms ACO, ACO-K-Means, ACO-SLC,
ACO-SLC-LWCR, ACO-SLC-Mixture are tested and compared. In the following
simulation, ACO refers to Ant-cycle presented by Dorigo, which is very
typical \cite{A. Colorni11}.

All test data in this paper is downloaded from http://www.iwr.
uniheidelberg.de /iwr/comopt/soft /TSPLIB95/TSPLIB.html. All algorithms in
this paper run on personal computer. CPU: 1.80GHz. Memory: 480M. Software:
Matlab. The parameters are listed as below. Initialize pheromone trails $%
\tau _{ij}(0)=1$, iteration number $1000$, $\varepsilon =0.001$, $\alpha =1$%
, $\beta =10$, $\rho =0.4$, $Q=300$, $m=[\frac{N}{1.5}]$. Two performance
items are tested. One item is the running time, which is defined as $%
Ratio=Time(ACO)/Time(Algorithm)$. The bigger the ratio is, the faster the
algorithm is. The other item is the quality of solution, which is defined as
the percentage of error $Error=(Solution-Optimum)/Optimum$, where $Optimum$
denotes the best solution known currently. The smaller the error is, the
better the quality of solution is.

\begin{figure}[tbh]
\begin{center}
\epsfig{file=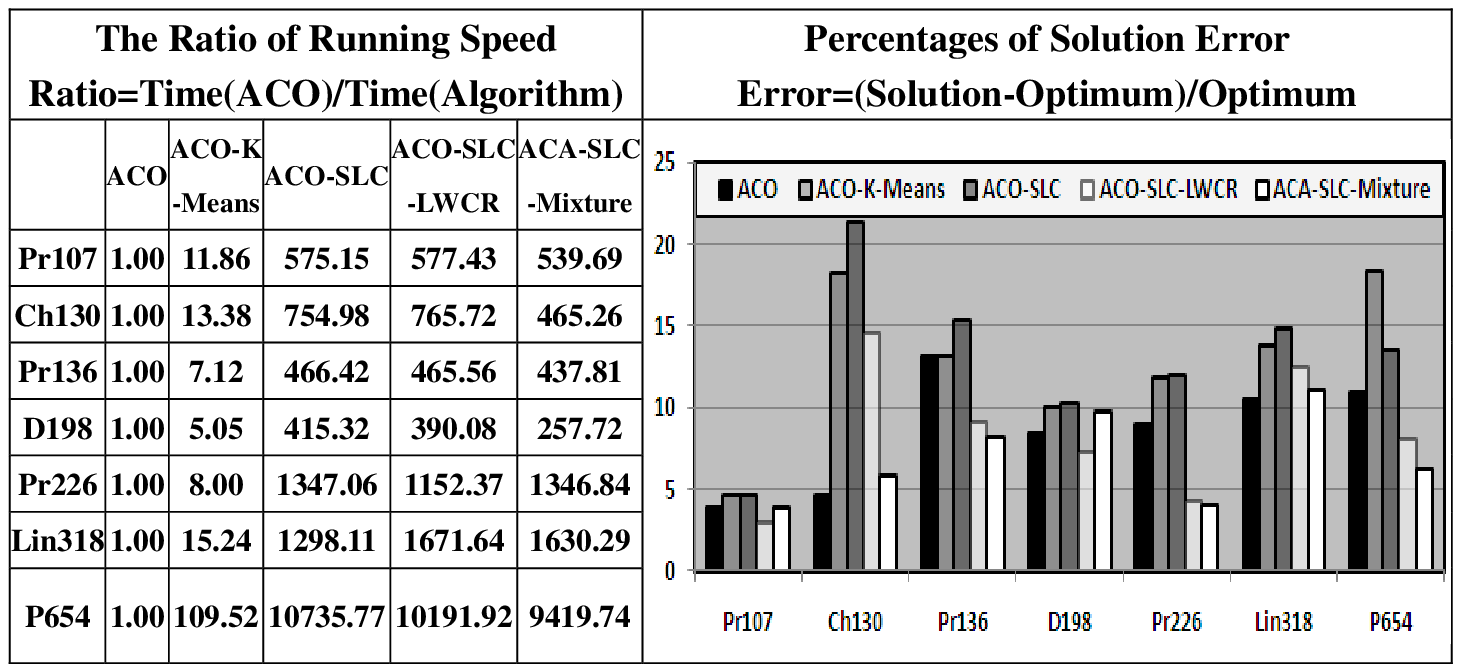,width=13cm,}
\end{center}
\caption{The Performance Comparison of The Five Algorithms of ACO,
ACO-K-Means, ACO-SLC ACO-SLC-LWCR and ACO-SLC-Mixture: The figure shows that
ACO-SLC algorithm, ACO-SLC, ACO-SLC-LWCR and ACO-SLC-Mixture are faster than
ACO by 415\symbol{126}10736, 390\symbol{126}10192 and 257-9419 of factors
respectively! However, some solutions of ACO-K-Means and ACO-SLC have low
quality. ACO-SLC-Mixture can process mixture distribution and its inaccuracy
ratio is less than ACO in most cases, and is bigger than ACO by 2\% at most.
}
\label{fig5}
\end{figure}

The performances of the five algorithms are listed in Fig.\ref{fig5}. It
shows that ACO-SLC, ACO-SLC-LWCR and ACO-SLC-Mixture are faster than ACO by
415\symbol{126}10736, 390\symbol{126}10192 and 257-9419 of factors
respectively! However, some solutions of ACO-K-Means and ACO-SLC have low
quality. The inaccuracy ratio of ACO-SLC-Mixture is less than ACO in most
cases, and is bigger than ACO by 2\% at most.

\textbf{The Defect of ACO-SLC:} The authors have done many simulations under
different conditions. And these simulations show the quality of ACO-SLC
solution depends on the quality of clustering and clustering quality of SLC
is sensitive to the initial centroids just liking K-Mean algorithm. This is
main defect of ACO-SLC. All cities are two-dimensional (or
three-dimensional) points and they are visual for general TSP problems.
Hence, reasonable initial centroids can be selected by researcher's eyes to
reduce the sensitivity of initial centroids. The initial centroids of the
simulations of this paper is selected by the authors' eyes, and they are put
at appendix III of this paper.

\section{CONCLUSION}

\textbf{Time Complexity of ACO:} ACO is the algorithm that inspired by the
foraging behavior of ant colonies and has be applied to solve many
optimization problem. The typical application of ACO is the application at
Traveling Salesman Problem (TSP). The running time of ACO is $O(t_{\max
}MN^{2})$, where $t_{\max }$, $M$ and $N$ denotes the iteration number,
number of ants, and number of cities respectively. Parameter $m$ is a
experiential value and is set to $[\frac{N}{1.5}]$ in general. Parameter $N$
is the key factor of running time because running time is proportional to
its square. Parameter $t_{\max }$ and $N$ are available, and decreseaing
parameter $t_{\max }$ and $N$ will cut down running time.

\textbf{Focus of ACO Study:} ACO can generate solution with high quality in
general. But its shortage is that running time is too long. Cutting down
running time is one of study focuses of ACO, and one way is to decrease
parameter $t_{\max }$ and $N$, especially $N$.

\textbf{Basic Idea for This Study Focus:} For this study focus, the
following basic idea is presented in this paper.

Firstly, all cities are classified into compact classes, where compact class
is the class that all cities in this class cluster tightly in a small region.

Secondly, let ACO act on every class to get a local TSP route.

Thirdly, all local TSP routes are jointed to form solution.

Fourthly, the inaccuracy of solution caused by clustering is eliminated.

\textbf{Realization of Basic Idea:} The realization of above idea is based
on a novel clustering algorithm presented in this paper, which is named
Special Local Clustering algorithm (SLC). The running time of SLC is far
less than the time of ACO. SLC generates compact classes, while current
popular clustering algorithm such as K-Means does not generate compact
classes in general. The compactness of class makes the length of TSP route $%
L_{t}$ at every iteration convergent, the convergence of $L_{t}$ (i.e., $%
(\left\vert L_{t}-L_{t+1}\right\vert /L_{t})\rightarrow 0$) is proposed as
the termination criterion of ACO in this paper. Thus, parameter $t_{\max }$
is cut down to improve the running speed of ACO. In addition, every class
has small size, ACO acting on small class makes parameter $N$ is cut down,
and running speed is improved. According to this analysis, ACO-SLC algorithm
is presented in this paper. Simulation shows that ACO-SLC is faster than ACO
by 415\symbol{126}10736 of factors!

\textbf{Elimination of the Solution Inaccuracy Caused by Clustering:}
Although the running speed is improved in this paper, the inaccuracy of
solution is heavy. Two factors causing the inaccuracy are found in this
paper. One is the cross-edges (see section 3.3), and the other factor is the
unmatch between ACO-SLC and mixture distribution (see section 3.4).
According to these two factors, ACO-SLC-LWCR and ACO-SLC-Mixture\ are
presented in this paper, which is the improvement of ACO-SLC.\ Simulation
shows that ACO-SLC-LWCR and ACO-SLC-Mixture is faster than ACO by 390\symbol{%
126}10192 and 257-9419 of factors respectively! The inaccuracy ratio of
ACO-SLC-Mixture is less than ACO in most cases, and is bigger than ACO by
2\% at most.

\section{ACKNOWLEDGMENT}

The authors appreciate the help from Prof. J. Zhang, Prof. J. Zhou and Prof.
Qi Li. The authors would like to thank group members Mr. C.-B. Wang and Ms.
Q. Yang for their check.

\section{APPENDIX}

\textbf{Appendix I:} Chain-Shaped Clustering Algorithm of Ref.\cite{HUANG
Xiaobin53}

{\tiny This algorithm is used to process the data set which distribution is
chain-shaped only, and is introduced as below.}

{\tiny Step1. Unify all data and calculate the centroid of data set, and
select the point that is farthest from the centroid as the seed of the first
class.}

{\tiny Step2. Select the points that are close to the seed into the first
class as many as possible until the trace of the first set is bigger than
the pre-assigned threshold (Notice: The set of random data has covariance
matrix. The trace of covariance matrix is proportion to the size of region
containing these random data. The smaller the trace is, the smaller the size
is. Thus, if the trace is less than a pre-assigned threshold, the class is
still small and compact. The threshold is 0.0005 in this paper).}

{\tiny Repeating the two steps, the first class, second class, 3rd class,
and so on, will be obtained until all points are classified. Finally, merge
all neighbor classes as a new big class. E.g., if class A is close to B and
B close to C\ the three classes A, B and C are neighbors and are merged as a
big chain-shaped class.}

\textbf{Appendix II:} The Parameter Configuration of Little-Window Method

{\tiny The configuration of parameter $w$ in little-window strategy: Let $n$
denotes the number of neighbor cities, If $n<21$, let w=min($n$-1, 8); Else
if $n<101$, let w=min($n$-1, 9); Else if $n<144$, let w=min($n$-1,13); Else
if $n<1000$, let w=min($n$-1, 19); Else $ifn<4000$, let w=min($n$-1,100);
Else let $w=[n/10]$; }

\textbf{Appendix III:} Initial Centroids of Classes

\begin{tabular}{l}
{\tiny The initial centroids of classes which is used by clustering
algorithm SLC: Each initial centroids has coordinate (x, y)} \\
\begin{tabular}{ll}
{\tiny Pr136} &
\begin{tabular}{ll}
{\small x=} & {\tiny 5000 5000 5000 5000 \ 12000 12000 12000 12000} \\
{\small y=} & {\tiny 3000 5000 8000 10000 10000 8000 \ 5000 \ 3000}%
\end{tabular}
\\
{\tiny Pr226} &
\begin{tabular}{ll}
{\small x=} & {\tiny 550 \ 4800 4800 2000 4000 \ 12000 12500 10000 12500
16500 13000 8800} \\
{\small y=} & {\tiny 2000 7000 9000 9000 12000 12000 8500 \ 8500 \ 7000 \
3500 \ 2000 \ 3500}%
\end{tabular}
\\
{\tiny Ch130} &
\begin{tabular}{ll}
{\small x=} & {\tiny 50 90 \ 90 \ 40 \ 40 \ 150 300 200 270 200 260 400 400
450 550 400 560 700 700 700} \\
{\small y=} & {\tiny 50 190 340 450 680 100 50 \ 250 400 540 650 50 \ 150
300 200 540 400 200 380 550}%
\end{tabular}
\\
{\tiny Lin318} &
\begin{tabular}{ll}
{\small x=} & {\tiny 500 500 \ 500 \ 500 \ 1400 1400 1400 1100 2300 2300
2300 2000 3000 3000 3000 \ 3000} \\
{\small y=} & {\tiny 500 1500 3100 4000 500 \ 1800 3200 4000 500 \ 1700 3100
4000 500 \ 1500 \ 2700 3600}%
\end{tabular}
\\
{\tiny D198} &
\begin{tabular}{ll}
{\small x=} & {\tiny 3700 3670 3830 3800 3700 4000 2100 1800 1600 1300 950 \
700 \ 0 1500 1160 1500 1600 2100} \\
{\small y=} & {\tiny 1400 1400 1400 1150 1050 1050 1870 1600 1480 1400 1350
1000 0 1000 1170 1190 1400 1400}%
\end{tabular}
\\
{\tiny P654} &
\begin{tabular}{ll}
{\small x=} & {\tiny 1070 1070 1070 1500 1070 1070 1070 1070 2000 2500 2000
3500 4300 4500 5840 5840 5840 5840 5840 5840} \\
{\small y=} & {\tiny 2400 2200 2710 3600 4800 4500 5050 5200 2500 4000 5000
3500 3500 2400 2200 2710 2700 4900 4600 5300}%
\end{tabular}
\\
{\tiny Pr107} &
\begin{tabular}{ll}
{\small x=} & {\tiny 8500 8500 8500 8500 8500 8500 8500 8500 \ 8500 \ 16000
16000 16000 16000 16000 16000 16000 16000 16000} \\
{\small y=} & {\tiny 4800 5500 6400 7100 8000 8900 9800 10600 11500 4800 \
5500 \ 6400 \ 7100 \ 8000 \ 8900 \ 9800 \ 10600 11500}%
\end{tabular}%
\end{tabular}%
\end{tabular}

\end{document}